\documentclass{article}

\usepackage[accepted]{icml2019}
\usepackage[noend]{algpseudocode}
\usepackage{algorithm}

\usepackage[utf8]{inputenc} 
\usepackage[T1]{fontenc}    
\usepackage{hyperref}       
\usepackage{url}            
\usepackage{booktabs}       
\usepackage{amsfonts}       
\usepackage{nicefrac}       
\usepackage{microtype}      
\usepackage{comment}
\usepackage{subcaption}
\usepackage{todonotes}
\usepackage{graphicx}
\usepackage{setspace}
\usepackage{mathtools}
\usepackage{wrapfig}
\DeclareMathOperator*{\argmin}{arg\,min}
\DeclareMathOperator*{\argmax}{arg\,max}

\usepackage[toc,page]{appendix}

\icmltitlerunning{Imitating Latent Policies from Observation}
\begin{document}

\twocolumn[
\icmltitle{Imitating Latent Policies from Observation}

\begin{icmlauthorlist}
\icmlauthor{Ashley D. Edwards}{to}
\icmlauthor{Himanshu Sahni}{to}
\icmlauthor{Yannick Schroecker}{to}
\icmlauthor{Charles L. Isbell}{to}
\end{icmlauthorlist}

\icmlaffiliation{to}{Georgia Institute of Technology, Atlanta, GA, USA}

\icmlcorrespondingauthor{Ashley D. Edwards}{aedwards8@gatech.edu}

\icmlkeywords{reinforcement learning, state prediction, imitation learning, imitation from observation}

\vskip 0.3in
]
\printAffiliationsAndNotice{} 

\begin{abstract}
In this paper, we describe a novel approach to imitation learning that infers latent policies directly from state observations. We introduce a method that characterizes the causal effects of latent actions on observations while simultaneously predicting their likelihood. We then outline an action alignment procedure that leverages a small amount of environment interactions to determine a mapping between the latent and real-world actions. We show that this corrected labeling can be used for imitating the observed behavior, even though no expert actions are given. We evaluate our approach within classic control environments and a platform game and demonstrate that it performs better than standard approaches. Code for this work is available at \url{https://github.com/ashedwards/ILPO}.
\end{abstract}

\section{Introduction}
Humans often learn from and develop experiences through mimicry. Notably, we are capable of mirroring behavior through only the observation of state trajectories without direct access to the underlying actions ({\it e.g.}, the exact kinematic forces) and intentions that yielded them~\cite{rizzolatti2010functional}. In order to be general, artificial agents should also be equipped with the ability to quickly solve problems after observing the solution; however, imitation learning approaches typically require both observations and actions to learn policies along with extensive interaction with the environment.

A recent approach for overcoming these issues is to learn an initial self-supervised model for~\emph{how} to imitate by collecting experiences within the
environment and then using this learned model to infer policies from expert observations~\cite{pathak2018zero, torabi2018behavioral}. However, unguided exploration can be risky in many real-world scenarios and costly to obtain. Thus, we need a mechanism for learning policies from observation alone without requiring
access to expert actions {\it and} with only a few interactions within the environment. 

In order to tackle this challenge, we hypothesize that predictable, though unknown, causes may describe the classes of transitions that we observe. These
causes could be natural phenomena in the world, or the consequences of the actions that the agent takes. This work aims
to demonstrate how an agent can predict and then imitate these latent causes, even though the ground truth environmental
actions are unknown. 

We follow a two-step approach, where the agent first learns a policy offline in a latent space that best
describes the observed transitions. Then it takes a limited number of steps in the environment to ground this latent policy to the true action labels. We liken this to learning to play a video game by observing a friend play first, and then attempting to play it ourselves. By observing, we can learn the goal of the game and the types of actions we should be taking, but some interaction may be required to learn the correct mapping of controls on the joystick.

We first make the assumption that the transitions between states can be described through a discrete set of latent actions. We then learn a forward dynamics model that, given a state and latent action, predicts the next state and prior, supervised only by \{state, next state\} pairs. We use this model to greedily select the latent action that leads to the most probable next state. Because these latent actions are initially mislabeled, we use a few interactions with the environment to learn a relabeling that outputs the probability of the true action. 

We evaluate our approach in four environments: classic control with cartpole, acrobot, and mountain car, and a recent platform game by OpenAI, CoinRun~\cite{cobbe2018quantifying}. We show that our approach is able to perform as well as the expert after just a few steps of interacting with the environment, and performs better than a recent approach for imitating from observations, Behavioral Cloning from Observation~\cite{torabi2018behavioral}.

 \section{Related work}
Imitation learning approaches aim to train artificial and real-world agents to imitate expert behavior by providing a set of expert demonstrations. This approach has an extensive breadth of applications, ranging from early successes in autonomous
driving~\cite{pomerleau1989alvinn}, to applications in robotics~\cite{Schaal1997,chernova_robot_2014} and software agents (e.g.~\cite{silver2016mastering,nair2017overcoming}).
However, traditional approaches typically assume that the expert's actions are known. This often requires the data to be
specifically recorded for the purpose of imitation learning and drastically reduces the amount of data that is readily
available. Recent approaches that do not require expert actions typically must \textit{first} learn behaviors in the agent's environment through extensive interactions. Our approach first learns latent behaviors from the demonstration data only, followed by only a few necessary interactions with the environment. We now describe classic approaches to imitation learning along with more modern approaches.

\subsection{Classic approaches}
Arguably, the most straight-forward approach to imitation learning is behavioral cloning~\cite{pomerleau1989alvinn}, which
treats imitation learning as a supervised learning problem. More
sophisticated methods achieve better performance by reasoning about the state-transitions explicitly, but often require
extensive information about the effects of the agent's actions on the environment. This information can come either in the
form of a full, often unknown, dynamics model, or through numerous interactions with the environment. Inverse Reinforcement Learning (IRL) achieves this by using the demonstrated state-action pairs to
explicitly derive the expert's intent in the form of a reward function~\cite{Ng2000,abbeel2004apprenticeship}. 

\subsection{Direct policy optimization methods}
Recently, more direct approaches have been introduced that aim to match the state-action visitation frequencies observed by the
agent to those seen in demonstrations. GAIL~\cite{ho2016generative} learns to
imitate policies from demonstrations and uses adversarial training to distinguish if a state-action pair comes
from following the agent or expert's policy while simultaneously minimizing the difference between the two.
SAIL~\cite{NIPS2017_6884} achieves a similar goal by using temporal difference learning to estimate the gradient of
the normalized state-action visitation frequency directly. However, while these approaches are efficient in the amount
of expert data necessary for training, they typically require a substantial amount of interactions within the
environment.

\subsection{Learning from state observations}
Increasingly, works have aspired to learn from observation alone without utilizing expert actions. Imitation from Observation~\cite{liu2017imitation}, for example, learns to imitate from videos without actions and translates from one context to another. However, this approach requires using learned features to compute rewards for reinforcement learning, which will thus require many environment samples to learn a policy. Similarly, time-contrastive networks~\cite{sermanet2017time} and unsupervised perceptual rewards~\cite{sermanet2016unsupervised} train robots to imitate from demonstrations of humans performing tasks, and recent work used audio to align different YouTube videos to train an agent to learn Montezuma's revenge and Pitfall~\cite{aytar2018playing}. But these approaches also learn features for a reward signal that is later used for reinforcement learning.  Finally, both third-person imitation learning~\cite{stadie2017third} and GAIfO~\cite{torabi2018generative} extend GAIL for use with demonstration data that lacks actions, but these approaches also utilize a reward signal in a similar manner as GAIL. Therefore, while each of these approaches learn policies from state observations, they require an intermediary step of using a reward signal, whereas we learn the policy directly without performing reinforcement learning.

A recent approach aimed to learn from observations by first learning how to imitate in a self-supervised manner, then given a task, attempt it zero-shot~\cite{pathak2018zero}. However, this approach requires learning in the agent's environment first rather than initially learning from the observations. Another approach utilizes learned inverse dynamics to train agents from observation \cite{torabi2018behavioral}. A problem with such an approach is that learning a dynamics model usually requires a substantial number of interactions with the environment. Our work aims to first learn policies from demonstrations~\emph{offline}, and then only use a few interactions with the environment to learn the true action labels. 
 \begin{figure*}[htb]
 \centering
  \begin{subfigure}{.47\linewidth}
  	\centering
     \includegraphics[width=.6\linewidth]{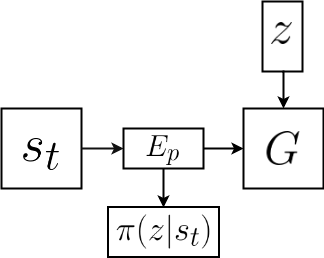}
     \caption{Latent Policy Network}
  \end{subfigure}
    \begin{subfigure}{.47\linewidth}
    \centering
     \includegraphics[width=.8\linewidth]{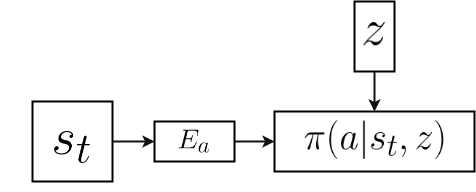}
     \caption{Action Remapping Network}
  \end{subfigure}%
     \caption{The latent policy network learns a latent policy, $\pi(z|s)$, and a forward dynamics model, $G$. The action remapping network learns $\pi(a|s_t,z)$ to align the latent actions $z$ with ground-truth actions $a$. We train embeddings, $E_a$ and $E_p$, concurrently with each network.}
       \label{fig:networks}
\end{figure*}
\subsection{Multi-modal predictions}
Our approach predicts forward dynamics given a state and latent action. This is similar to recent works that have learned action-conditional predictions for reinforcement learning environments~\cite{oh2015action, chiappa2017recurrent}, but those approaches utilize ground truth action labels. Rather, our approach learns a latent multi-modal distribution over future predictions. Other related works have utilized latent information to make multi-modal predictions. For example, BicycleGAN~\cite{zhu2017toward} learns to predict a distribution over image-to-image translations, where the modes are sampled given a latent vector. InfoGAN uses latent codes for learning interpretable representations~\cite{chen2016infogan}, and InfoGAIL~\cite{li2017infogail} uses that approach to capture latent factors of variation between different
demonstrations. These works, however, do not attempt to learn direct priors over the modes, which is crucial in our formulation for deriving policies.

As such, our approach is more analogous to online clustering, as it predicts multiple expected next states and priors over them. However, we do not have direct access to the clusters or means. Other works have aimed to cluster demonstrations, but these approaches have traditionally segmented different types of trajectories, which represent distinct preferences, rather than next-state predictions~\cite{hausman2017multi, babes2011apprenticeship}.

\section{Approach}
We now describe our approach, Imitating Latent Policies from Observation (ILPO), where we train an agent to imitate behaviors from expert state observations. 

 \begin{algorithm*}[htb]
\begin{algorithmic}[1]
	\Function{ILPO}{$s^*_0$, $s^*_1$, \ldots, $s^*_N$}
        \State \textit{Step 1: Learning latent policies}
    	\For{$k \gets 0 \ldots \text{\#\it{Epochs}}$} 
        	\For{$i \gets 0 \ldots {N-1}$} \Comment{\emph{(Omitting batching for clarity)}}
                \State Train latent dynamics parameters $\theta \gets \theta - \nabla_\theta \min_z \Vert G_{\theta}(E_{p\theta}(s^*_i), z) - s^*_{i+1}\Vert^2_2$
                \State Train latent policy parameters $\omega \gets \omega - \nabla_\omega \Vert \sum_z \pi_\omega(z|s^*_i) G_{\theta}(E_{p{\theta}}(s^*_i), z) - s^*_{i+1}) \Vert^2_2$
            \EndFor
        \EndFor
        \Statex
        \State \textit{Step 2: Action remapping}
        \State Observe state $s_0$
        \For{$t \gets 0 \dots \text{\#\it{Interactions}}$}
        	\State Choose latent action $z_t\gets \argmax_z \pi_\omega(z|E_{a\xi}(s_t))$
            \State Take $\epsilon$-greedy action $a_t\gets \argmax_a \pi_\xi(a|z_t, E_{a\xi}(s_t))$
            \State Observe state $s_{t+1}$
            \State Infer closest latent action $z_t=\argmin_{z}\Vert E_{p\theta}(s_{t+1}) - E_{p\theta}(G_\theta(E_{p\theta}(s_t), z)) \Vert_2$
            \State Train action remapping parameters $\xi \gets \xi + \nabla_\xi \log \frac{\pi_\xi(a_t|z_t,E_{a\xi}(s_t))}{\sum_a \pi_\xi(a|z_t,E_{a\xi}(s_t))}$
            
        \EndFor
    \caption{Imitating Latent Policies from Observation}
    \EndFunction
\label{algo:ilpo}
\end{algorithmic}
\end{algorithm*}
\subsection{Problem formulation}
\label{sec:formulation}
We aim to use ILPO to solve problems specified through a Markov Decision Process (MDP) <$S, A, R, T$>~\cite{sutton1998reinforcement}. Here, $s \in S$ denotes the states in the environment, $a_t \in A$ corresponds to actions, $r_t \in R$ are the rewards the agent receives in each state, and $T(s_t, a_t, s_{t+1})$ is the transition model, which we assume is unknown. Reinforcement learning approaches aim to learn policies $\pi(a|s_t)$ that determine the probability of taking an action $a$ in some state $s_t$. We use imitation learning to directly learn the policy and use the reward $r_t$ only for evaluation purposes. 

We are given a set of expert demonstrations described through noisy state observations $\{s^*_1 \dots s^*_n\} \in D$. In our approach, we will use these observations to predict a multimodal forward dynamics model. As such, noise is necessary for ensuring that state transitions are properly modeled.

Given two consecutive observations $\{s_t, s_{t+1}\}$, we define $z$ as a~\emph{latent action} that caused this transition to occur.  As such, the action spaces that we consider are discrete with deterministic transitions.  Because our problems are specified through MDPs, we assume that the number of actions, $|A|$, is known. Hence, we can define $\{z_1 \dots z_{|A|}\} \in Z$ latent actions, where $|Z|=|A|$ is used as an initial guess for the number of latent actions. However, there may be more or less types of transitions that appear in the demonstration data. For example, if an agent has an action to move left but always moves right, then the "left" transition will not be observed. Or if the agent moves right and bumps into a wall, this stationary transition may appear to be another type of action. As such, we will empirically study the effect of using latent actions when $|Z| \neq |A|$.

\subsection{Behavioral cloning}
Given expert states and actions $\{s_1, a_1 \dots s_n, a_n \}$, behavioral cloning uses supervised learning to approximate $\pi(a|s_t)$. That is, given a state $s_t$, this approach predicts the probability of taking each action, i.e., the policy. However, imitation by observation approaches do not have access to expert actions. To address this, behavioral cloning from observation (BCO)~\cite{torabi2018behavioral} first learns an inverse dynamics model $f(a|s_t, s_{t+1})$ by first collecting samples in the agent's environment. Then, the approach uses this model to label the expert observations and learn $\pi(a|s_t)$. However learning dynamics models online can require a large amount of data, especially in high-dimensional problems.

We make the observation that we do not need to know action labels to make an initial hypothesis of the policy. Rather, our approach learns a~\emph{latent} policy $\pi_\omega(z|s_t)$ that estimates the probability that a latent action $z$ would be taken when observing $s_t$. This process can be done~\emph{offline} and hence more efficiently utilizes the demonstration data. We then use a limited number of
interactions with the environment to learn an action-remapping network that efficiently associates the true actions the agent can take with the latent
policy identified by our learned model. 
\subsection{Step 1: Learning latent policies}\label{sec:step1}
In order to learn $\pi_\omega(z|s_t)$, we introduce a latent policy network with two key components: a latent forward dynamics model $G$ that learns to predict $\widehat{s}_{t+1}$, and a prior over $z$ given $s_t$, which gives us the latent policy, as shown in figure~\ref{fig:networks}. 

\subsubsection{Latent forward dynamics}
We first describe how to learn a latent forward dynamics model from expert state observations. Given an expert state $s_t$ and latent action $z$, our approach trains a generative model $G_\theta(E_p(s_t), z)$ to predict the next state $s_{t+1}$, where $E_p$ is an embedding that is trained concurrently. Similar to recent works that predict state dynamics~\cite{edwards2018forward, goyal2018recall}, our approach predicts the differences between states $\Delta_t = s_{t+1} - s_t$, rather than the absolute next state, and computes $s_{t+1} = s_t + \Delta_t$. 

When learning to predict forward dynamics, a single prediction, $f(s_{t+1}|s_t)$, will not account for the different modes of the distribution, i.e., the effects of each action, and will thus predict the mean over all transitions. When using an action-conditional model~\cite{oh2015action, chiappa2017recurrent}, learning each mode is straightforward, as we can simply make predictions based on the observed next state after taking each action, $f(s_{t+1}|s_t, a)$. However, in our approach, we do not know the ground truth actions that yielded a transition. Instead, our approach trains a generative model $G$ to make predictions based on each of the latent actions $z \in Z$, $f(s_{t+1}|s_t, z)$. 

To train $G$, we compute the loss as:
\begin{align} 
\mathcal{L}_{min} = \min_z \Vert \Delta_t - G_\theta(E_p(s_t), z)\Vert^2.
\end{align} 
To allow predictions to converge to the different modes, we only penalize the one closest to the true next observation, $s_{t+1}$. Hence the generator must learn to predict the closest mode within the multi-modal distribution. This approach essentially allows each generator to learn~\emph{transition clusters} for each type of transition that is represented through $\Delta_t$. If we penalized each of the next state predictions simultaneously, the generator would learn to always predict the~\emph{expected} next state, rather than each distinct state observed after taking a latent action $z$.

We use $\Delta_t$ to better guide the generator to learn distinct transitions. For example, if we have an agent moving in discrete steps along the $x$-axis, then moving right would yield positive transitions $\Delta=1$ and moving left would yield negative transitions $\Delta=-1$. Our approach aims to train the generator to learn these different types of transitions.

Note that since $G$ is learning to predict $\Delta_t$, we will need to add each prediction to $s_t$ in order to obtain a prediction for $s_{t+1}$. For simplicity, in further discussion we will refer to $G$ directly as the predictions summed with the state input $s_t$. 
  \begin{figure*}[htb]
 \centering
  \begin{subfigure}{.33\linewidth}
  	\centering
    \includegraphics[width=\linewidth]{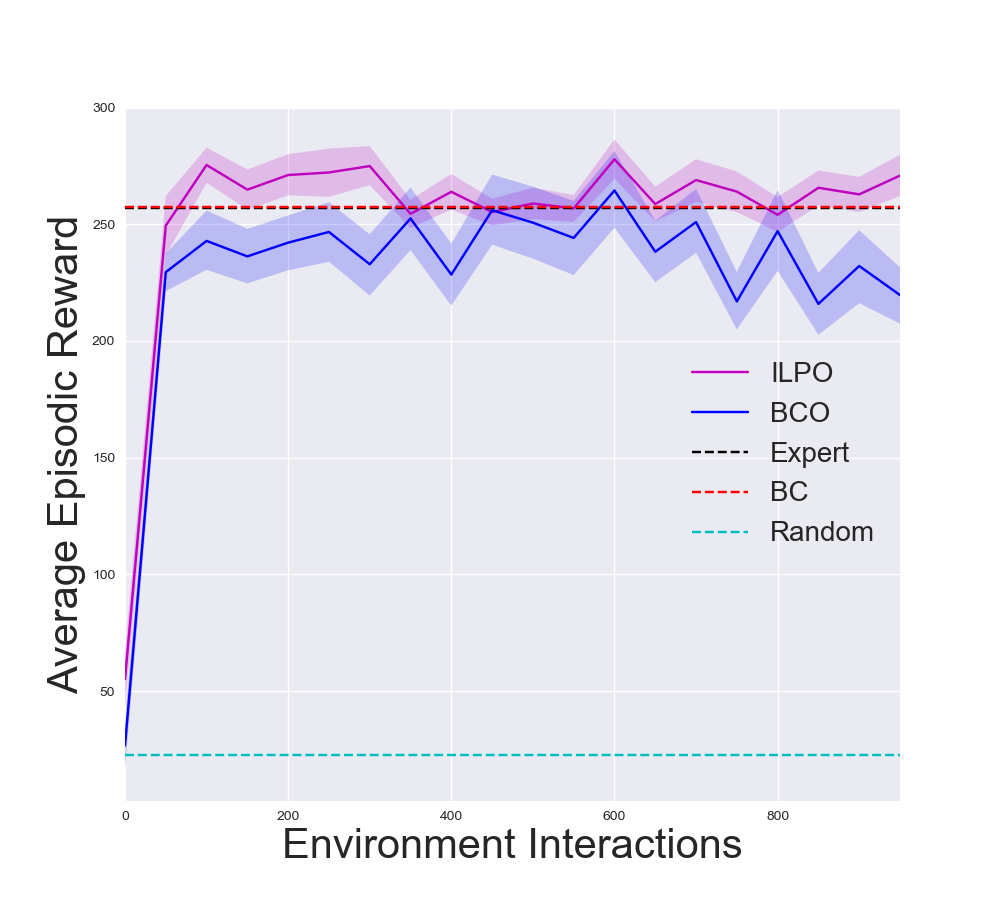}
    \caption{Cartpole}
  \end{subfigure}
    \begin{subfigure}{.33\linewidth}
    \centering
     \includegraphics[width=\linewidth]{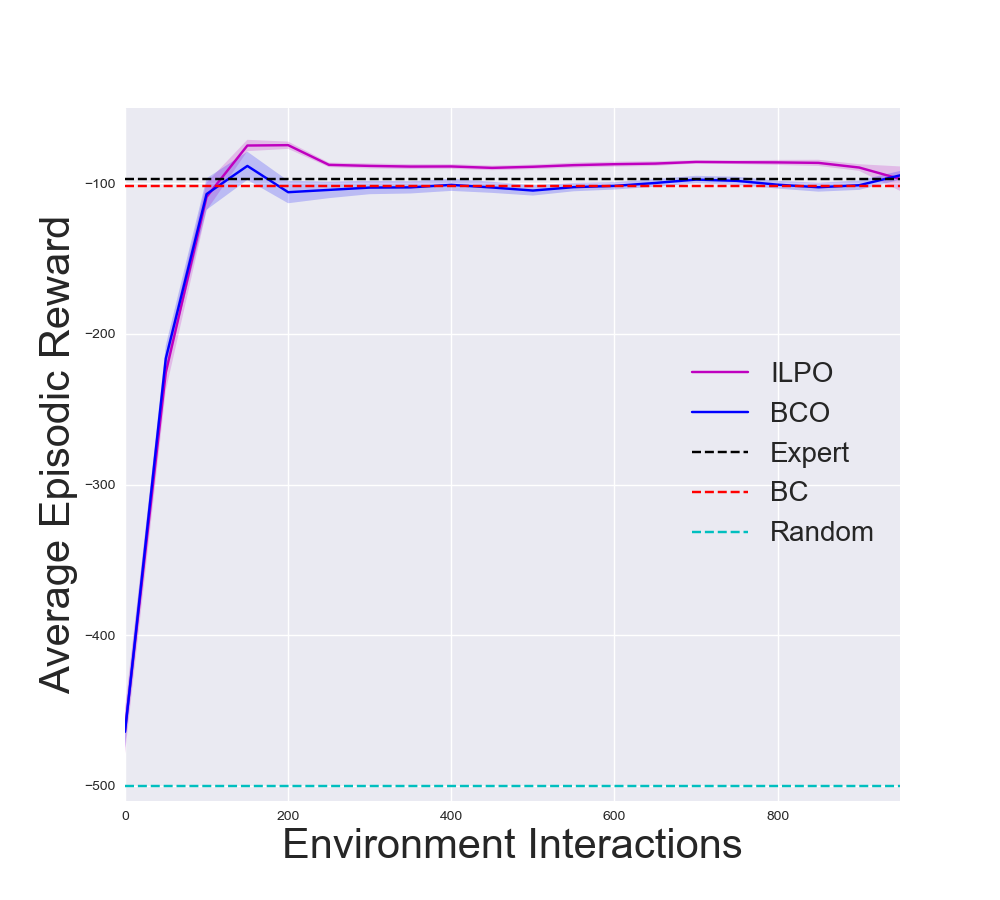}
     \caption{Acrobot}
  \end{subfigure}%
      \begin{subfigure}{.33\linewidth}
    \centering
     \includegraphics[width=\linewidth]{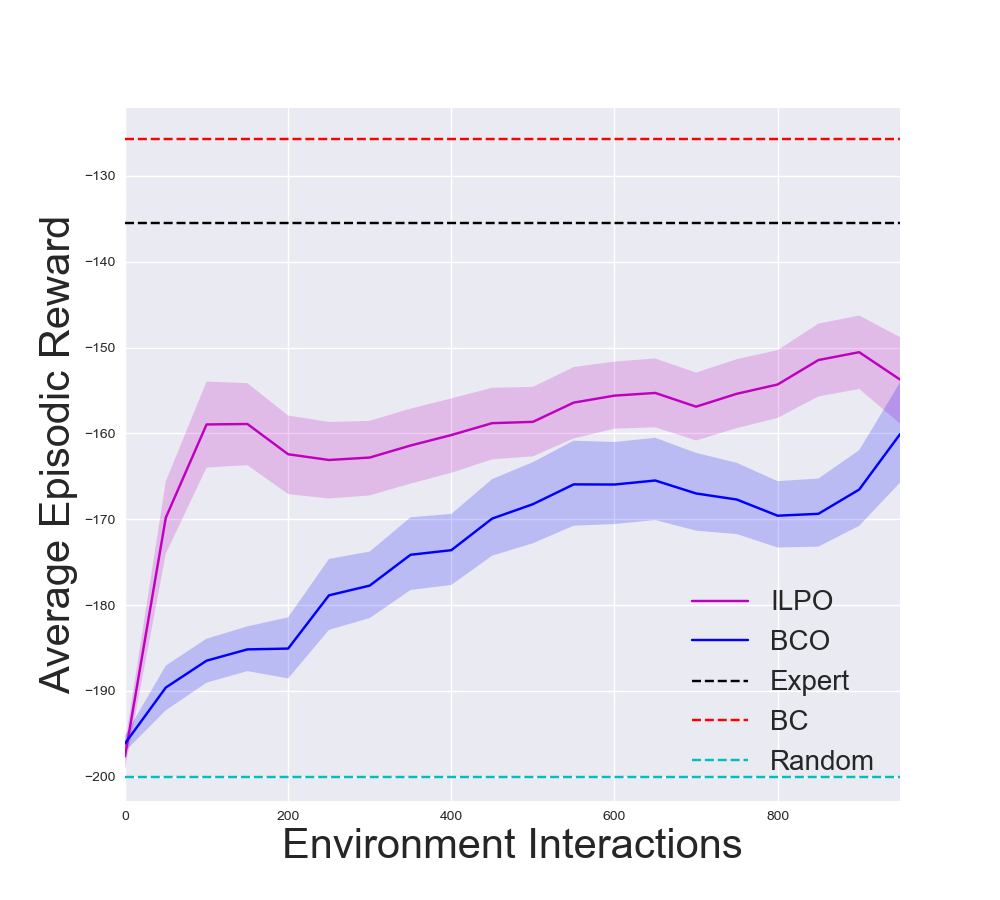}
     \caption{Mountain car}
  \end{subfigure}%
     \caption{Classic control imitation learning results. The trials were averaged over $50$ runs for ILPO and the policy was evaluated every $50$ steps and averaged out of $10$ policy runs. The reward used for training the expert and evaluation was +1 for every step that the pole was upright in cartpole, and a -1 step cost for acrobot and mountain car.}
  \label{fig:acrobot_results}
\end{figure*}
\subsubsection{Latent policy learning}
Crucially, ILPO concurrently learns the latent policy $\pi_\omega(z|E_p(s_t))$. This represents the probability that given a state $s_t$, a latent transition of the type $z$ will be observed in the expert data. We train this by computing the expectation of the generated predictions under this distribution, i.e., the expected next state, as:
\begin{align} 
\widehat{s}_{t+1} &= \mathop{\mathbb{E_{\pi\omega}}}[s_{t+1}|s_t]\\
                  &= \sum_z \pi_\omega(z|s_t)G_\theta(E_p(s_t), z).
\end{align}
We then minimize the loss as: 
\begin{align} 
\mathcal{L}_{exp} = \Vert s_{t+1} - \widehat{s}_{t+1} \Vert^2
\end{align} 
while holding the individual predictions fixed. This approach predicts the probability of each transition occurring. In other terms, this determines the most likely transition cluster.

With this loss, the latent policy is encouraged to make predictions that yield the most likely next state. For example, if an agent always moves right, then we should expect, given some state, for the next state to reflect the agent moving right. Any other type of transition should have a low probability so that it is not depicted within the next state.

The network is trained using the combined loss:
\begin{align} 
 \mathcal{L}_{policy} = \mathcal{L}_{min} + \mathcal{L}_{exp}.
\end{align}
We outline the training procedure for this step in lines 3-6 in algorithm~\ref{algo:ilpo}.

\subsection{Step 2: Action Remapping}\label{sec:step2}
In order to imitate from expert observations, the agent needs to learn a mapping from the latent policy learned in the previous step to the true action space: $\pi_\xi(a_t|z, E_a(s_t))$, where $E_a$ is an embedding that is trained concurrently. As such, it is invariantly necessary for the agent to explore the effect of its own actions within its environment. However, unlike BCO and other imitation from observation approaches, ILPO only needs to learn a mapping from $a$ to $z$ rather than a full dynamics model. The mapping $\pi_\xi$ also depends on the current state $s_t$ because latent actions are not necessarily invariant across states. The actions being predicted by each generator might change in different parts of the state-space. If an agent is flipped upside-down, for example, then the action "move up" would then look like "move down".

Nevertheless, generalization capabilities of neural networks should encourage a strong correlation between $a$ and $z$. The dynamics in two states are often more similar for the same action than they are for two different ones, thus assigning
the latent actions to the same type of transition should allow the network to generalize more easily. This intuition will allow us to learn such a mapping from only a few interactions with the environment, but is not a requirement for learning, and the algorithm will be able to learn to imitate the expert's policy regardless.

\subsubsection{Collecting experience}
To obtain training data for the remapped policy $\pi_\xi$, we allow the agent to interact with the environment to collect experiences in the form of $\{s_t, a_t, s_{t+1}\}$ triples. This interaction can follow any policy, such as a random policy or one that is updated in an online way. The only stipulation is that a diverse section of the state space is explored to facilitate generalization. We choose to iteratively refine the remapped policy $\pi_\xi$ and collect experiences by following this current estimate, in addition to an $\epsilon$-greedy exploration strategy.
  \begin{figure*}[htb]
 \centering
  \begin{subfigure}{.33\linewidth}
  	\centering
    \includegraphics[width=\linewidth]{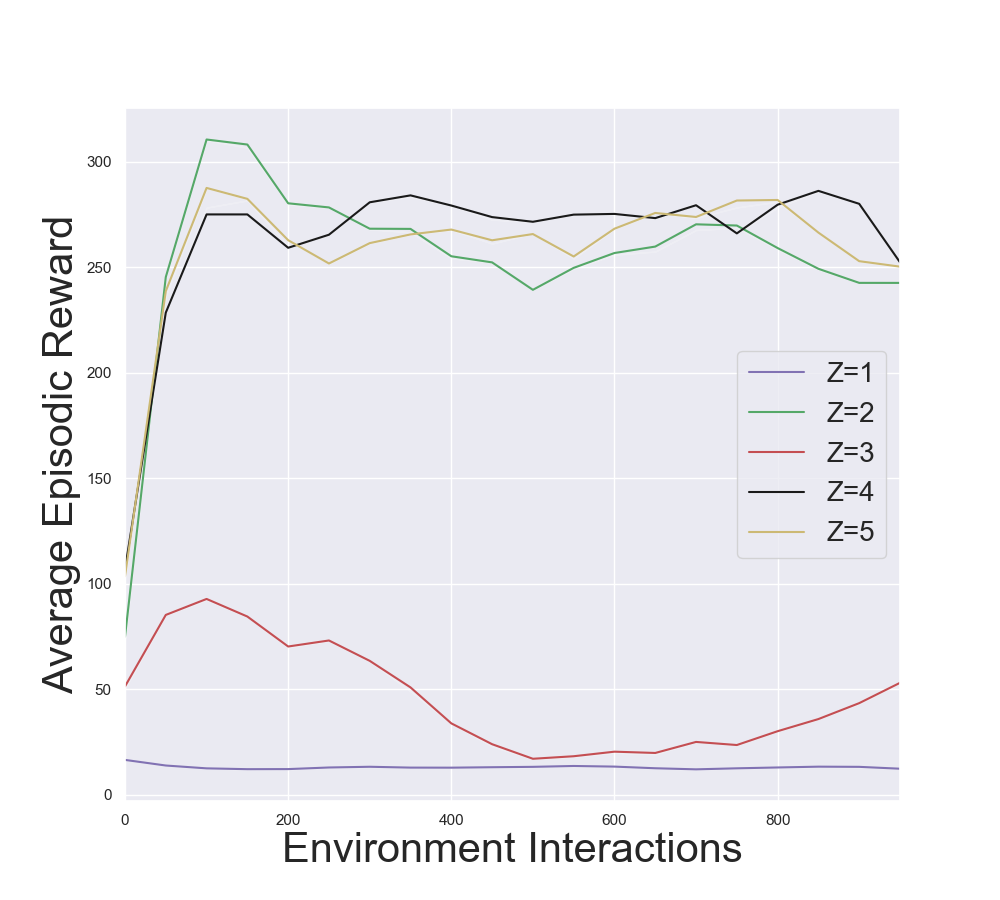}
    \caption{Cartpole}
  \end{subfigure}
    \begin{subfigure}{.33\linewidth}
    \centering
     \includegraphics[width=\linewidth]{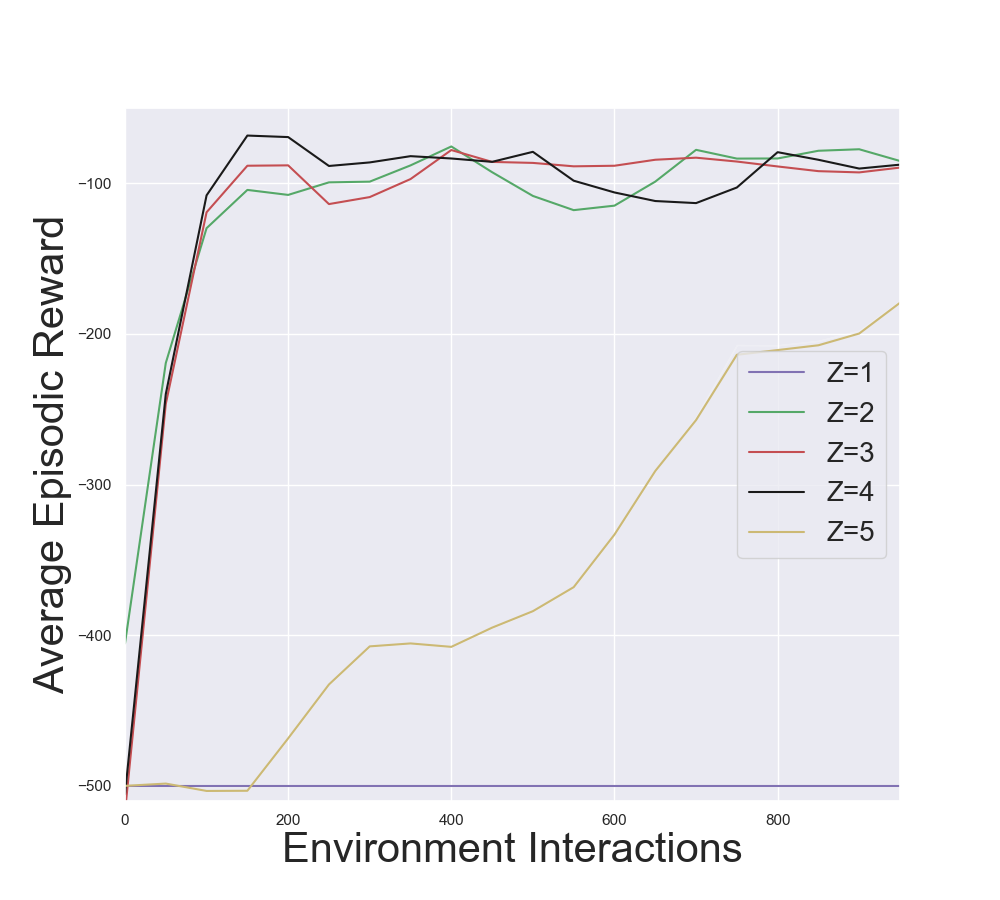}
     \caption{Acrobot}
  \end{subfigure}%
     \caption{Cartpole and Acrobot results for selecting $|Z|$. The trials were averaged over $5$ runs for ILPO and the policy was evaluated every $50$ steps and averaged out of $10$ policy runs. The reward used for training the expert and evaluation was +1 for every step that the pole was upright in cartpole, and a -1 step cost for acrobot.}
  \label{fig:acrobot_hyper}
\end{figure*}

\subsubsection{Aligning actions}
While collecting experiences $\{s_t, a, s_{t+1}\}$ in the agent's environment, we proceed in two steps to train the remapped policy. First, we identify the latent action that corresponds to the environmental state transitions $\{s_t, s_{t+1}\}$ and then we use the environmental action $a$ taken as a label to train $\pi_\xi(a_t|z_t, E_a(s_t))$ in a supervised manner. 

To do this, given state $s_t$, our method uses the latent dynamics model, $G$, trained in step 1, to predict each possible next state $\widehat{s}_{t+1}$ after taking a latent action $z$. Then it identifies the latent action that corresponds to the predicted next state that is the most similar to the observed next state $s_{t+1}$:
\begin{align}
    z_{t} = \argmin_z \Vert s_{t+1} - G_\theta(E_p(s_t), z) \Vert_2.
   \label{eq:distance}
\end{align}
To extend this approach to situations where euclidean distance is not meaningful (such as high-dimensional
visual domains), we may also measure distance in the space of the embedding $E_p$ learned in the previous step. In these domains, the latent action is thus given by:
\begin{align}
    z_{t} = \argmin_z \Vert E_p(s_{t+1}) - E_p\left(G_\theta(E_p(s_t), z)\right) \Vert_2.
    \label{eq:embedded_distance}
\end{align}
Having obtained the latent actions $z_t$ most closely corresponding to the environmental action $a_t$, we then train $\pi(a_t|z, s_t)$ as a
straight forward classification problem using a cross-entropy loss:
\begin{align}
    \mathcal{L}_{map} = \log \frac{\pi_\xi(a_t|z_t,E_{a}(s_t))}{\sum_a \pi_\xi(a|z_t,E_{a}(s_t))}.
\end{align}
\subsubsection{Imitating latent policies from observation}
Combining the two steps into a full imitation learning algorithm, given a state $s_t$, we use the latent policy outlined in step $1$ to identify the latent cause that is most likely to have the effect that the expert
intended, $z^*=\argmax_z \pi_\omega(z|s_t)$, and subsequently identify the action that is most likely to cause this effect, $a^*=\argmax_a \pi_\xi(a| z^*, s_t)$. The agent can then follow this policy to imitate the expert's behavior without having seen any expert actions. We outline the training procedure for this step in lines 7-14 in algorithm~\ref{algo:ilpo}.

 \section{Experiments and results}
In this section, we discuss the experiments used to evaluate ILPO. We aim to demonstrate that our approach is able to imitate from state observations only and with little interactions with the environment. In addition to this, we aim to show that learning dynamics online through environment interactions is less sample efficient than learning a latent policy first. In these experiments, we compare the environment interactions in ILPO during the action remapping phase with online data collection in BCO used for learning the inverse dynamics model. These samples are obtained after following the policies of each respective method, and we evaluate each approach after the same number of interactions.

We evaluate ILPO within classic control problems as well as a more complex visual domain. We used OpenAI Baselines~\cite{baselines} to obtain expert policies and generate demonstrations for each environment. We compare ILPO against this expert, a random policy, and Behavioral Cloning (BC), which is given ground truth actions, averaged over $50$ trials, and Behavioral Cloning from Observation (BCO). More details can be found in the appendix.

\subsection{Classic control environments}
  \begin{figure}[htb]
 \centering
 \begin{subfigure}{.3\linewidth}
  	\centering
    \includegraphics[height=2.3cm]{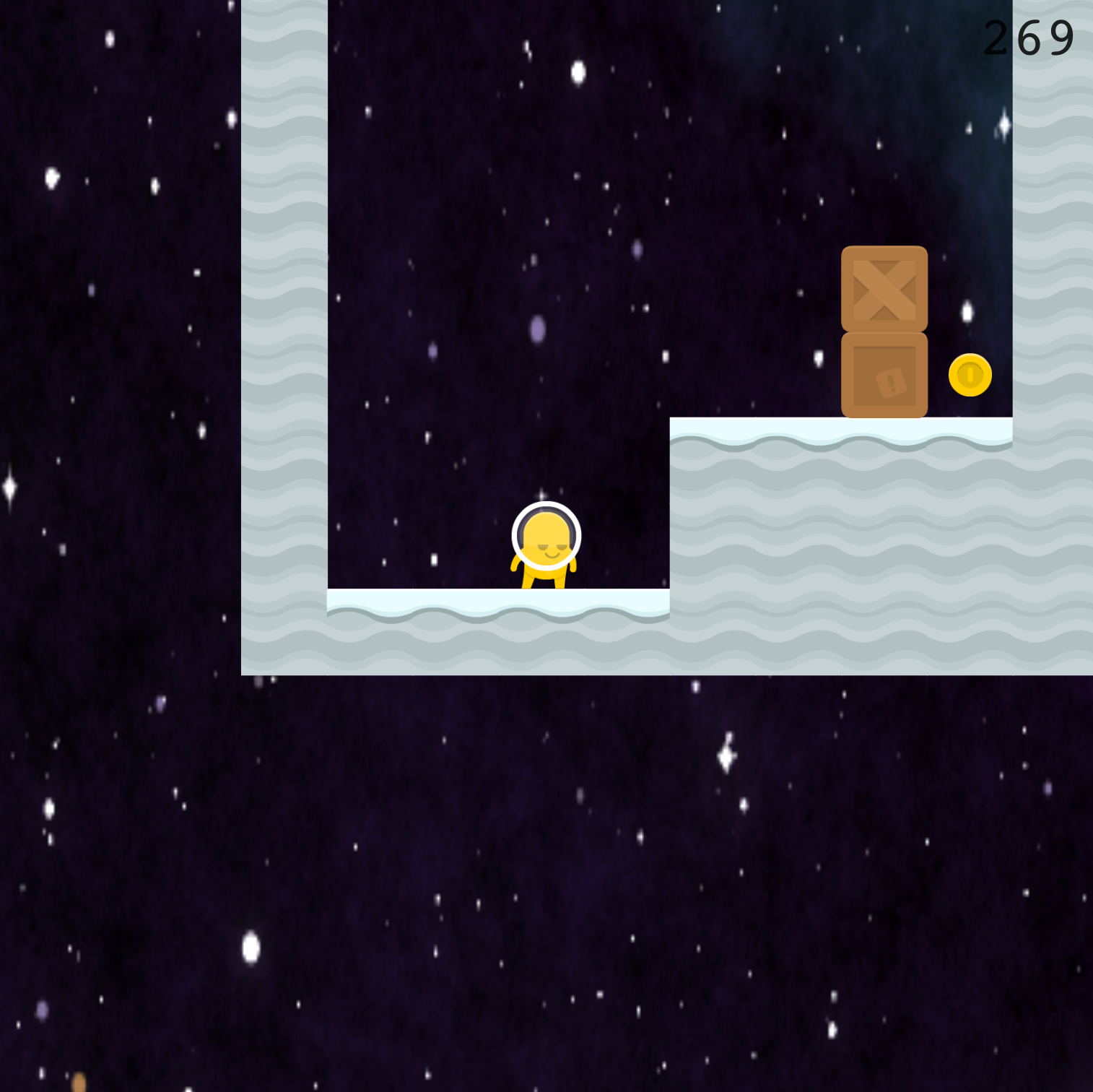}
  \end{subfigure}
  \begin{subfigure}{.3\linewidth}
  	\centering
    \includegraphics[height=2.3cm]{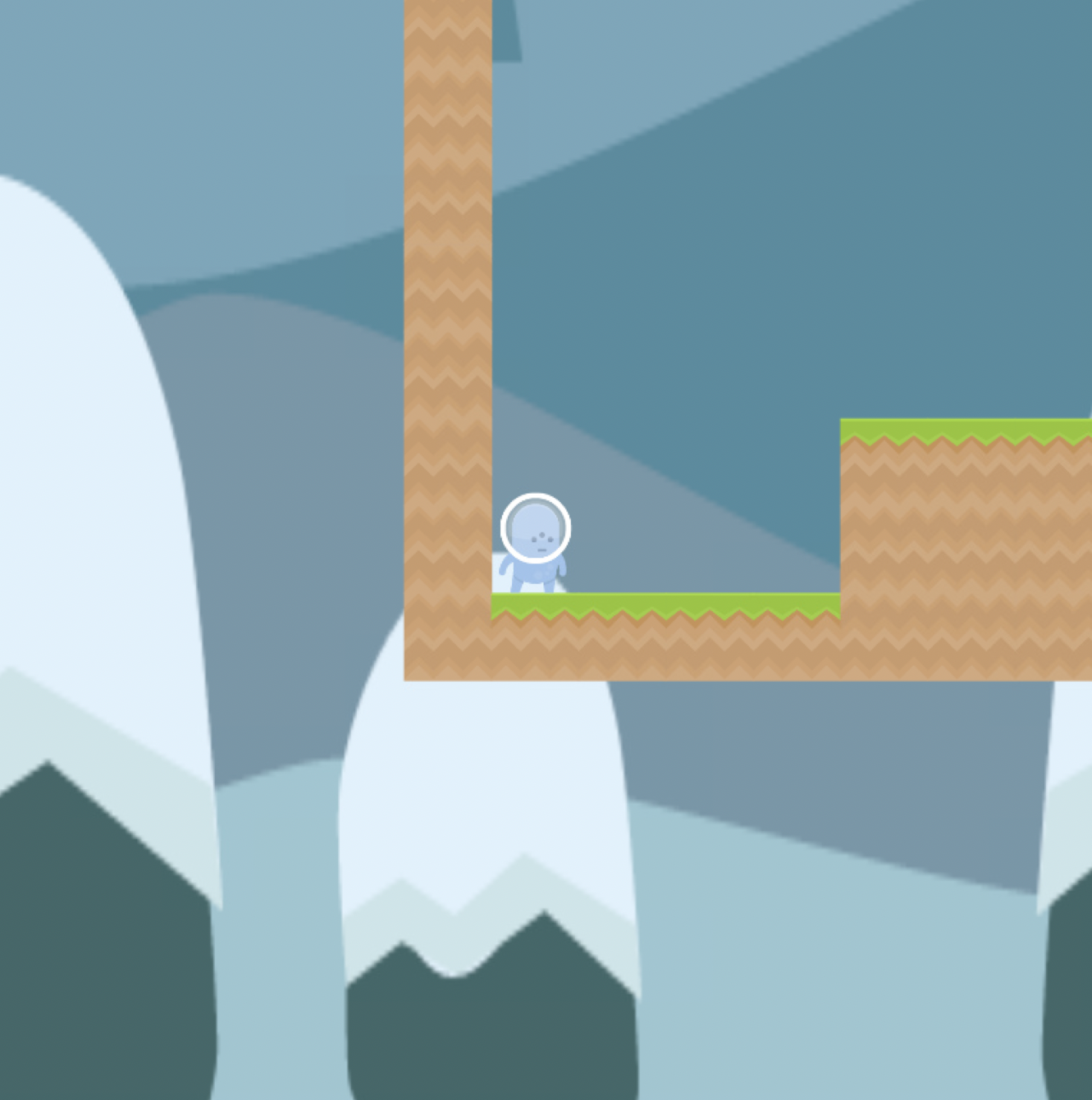}
  \end{subfigure}
    \begin{subfigure}{.3\linewidth}
    \centering
     \includegraphics[height=2.3cm]{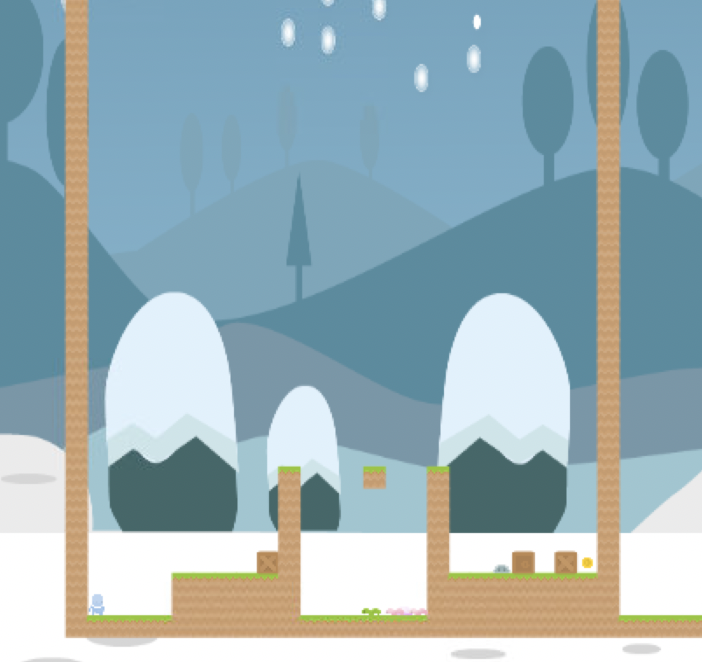}
  \end{subfigure}%
     \caption{CoinRun environment used in experiments. CoinRun consists of procedurally generated levels and the goal is to get a single coin at the end of a platform. We used an easy level (left) and hard level (middle and right) in our experiments. The middle image is the agent's state observation in the hard level and the image on the right is a zoomed out version of the environment. This task is difficult because the gap in the middle of the platform can not be recovered from and there is a trap once the agent reaches the other side. When training, the images also include a block showing x and y velocities.}
       \label{fig:thor}
\end{figure}
  \begin{figure}[htb]
 \centering
  \begin{subfigure}{.7\linewidth}
  	\centering
    \includegraphics[width=\linewidth]{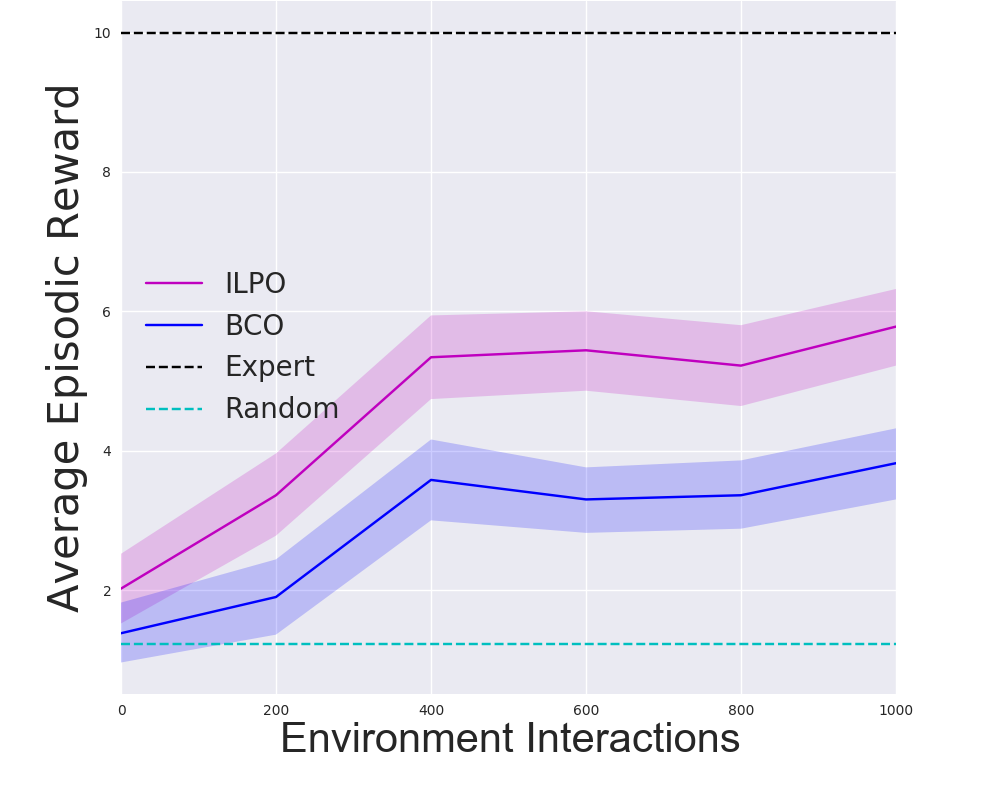}
    \caption{CoinRun easy}
  \end{subfigure}
    \begin{subfigure}{.7\linewidth}
    \centering
     \includegraphics[width=\linewidth]{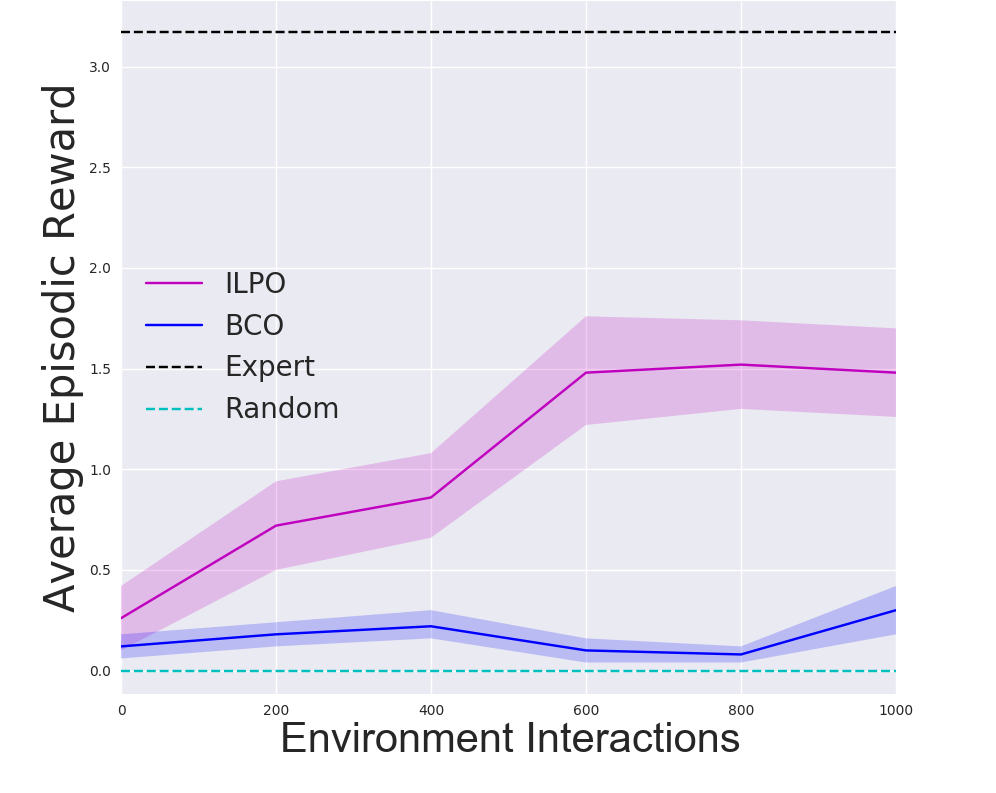}
     \caption{CoinRun hard}
  \end{subfigure}%
     \caption{CoinRun imitation learning results. The trials were averaged over $50$ runs for ILPO and BCO the policy was evaluated every $200$ steps and averaged out of $10$ policy runs. The reward used for training the expert and evaluation was $+10$ after reaching the coin. }
       \label{fig:thor_results}
\end{figure}
We first evaluated our approach within classic control environments~\cite{sutton1998reinforcement}. We used the standard distance metric from equation~\ref{eq:distance} to compute the distances between observed and predicted next states for ILPO. We used the same network structure and hyperparameters across both domains, as described in the appendix. We used $50,000$ expert state observations to train ILPO and BCO, and the corresponding actions to train Behavioral Cloning (BC). 
 
\textbf{Cartpole} is an environment where an agent must learn to balance a pole on a cart by applying forces of $-1$ and $1$ on it. The state space consists of $4$ dimensions: $\{x, \dot{x}, \theta, \dot{\theta}\}$, and the action space consists of the $2$ forces. As such, ILPO must predict $2$ latent actions and generate predicted next states with $4$ dimensions.
 
\textbf{Acrobot} is an environment where an agent with $2$ links must learn to swing its end-effector up by applying a torque of $-1$, $0$, or $1$ to its joint. The state space consists of $6$ dimensions: $\{\cos\theta_1, \sin\theta_1, \cos\theta_2, \sin\theta_2, \dot{\theta_1}, \dot{\theta_2}\}$, and the action space consists of the $3$ forces. As such, ILPO must predict $3$ latent actions and generate a predicted next state with $6$ dimensions. 

\textbf{Mountain car} is an environment where an agent on a single-dimension track must learn to push a car up a mountain by applying a force of $-1$, $0$, or $1$ to it. The state space consists of $2$ dimensions: $\{x, \cdot{x} \}$, and the action space consists of the $3$ forces. As such, ILPO must predict $3$ latent actions and generate a predicted next state with $2$ dimensions. 

 \subsubsection{Results}
 Figure~\ref{fig:acrobot_results} (left) shows the imitation learning results in cartpole. ILPO learns the correct policy and is able achieve the same performance as the expert and behavioral cloning in less than $100$ steps within the environment. Furthermore, ILPO performs much better than BCO, as it does not need to learn state-transitions while collecting experience, only a mapping from latent to real actions.
  
 Figure~\ref{fig:acrobot_results} (middle) shows the imitation learning results in acrobot. ILPO again learns the correct policy after a few steps and is able achieve as good of performance as the expert and behavioral cloning, again within $100$ steps. While BCO learns quickly, ILPO again performs better than it.
 
 Finally, the results for mountain car are shown in  Figure~\ref{fig:acrobot_results} (right). Neither ILPO nor BCO performed as well as the expert. Nevertheless, it is clear that ILPO outperforms BCO. 
 
 We were also interested in evaluating the effect of using a different number of latent actions. These results are shown in figure~\ref{fig:acrobot_hyper}. We see that choosing $|Z|=|A|$ is a good initial guess for the size of $Z$, but the agent is also able to learn from other sizes. $|Z|=1$ performed poorly in both acrobot and cartpole. This is because every action will collapse to a single latent and the state predictions cannot be disentangled.
 
 As we mentioned in section~\ref{sec:formulation}, ILPO requires stochastic demonstrations. We found that although the agent was capable of performing well with deterministic demonstrations, the performance decreased in this setting. See the appendix for more discussion.  

 \subsection{CoinRun}
 \begin{figure*}[htb]
  \begin{subfigure}{.13\linewidth}
  \centering
     \includegraphics[height=2.2cm]{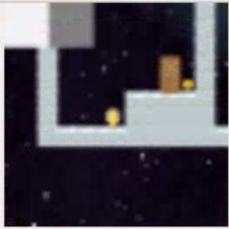}
     \caption{State}
  \end{subfigure}
    \begin{subfigure}{.2\linewidth}
    \centering
     \includegraphics[height=2.2cm]{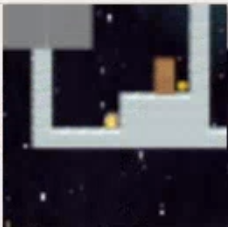}
     \caption{Next state}
  \end{subfigure}
    \begin{subfigure}{.35\linewidth}
    \centering
     \includegraphics[height=2.2cm]{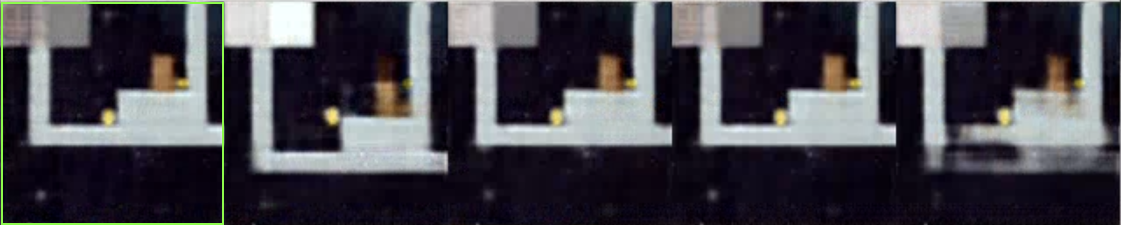}
     \caption{ILPO predictions}
  \end{subfigure}%
 \caption{Next state predictions computed by ILPO in the CoinRun easy task. The highlighted state represents the closest next state obtained from equation~\ref{eq:embedded_distance}.}
 \label{fig:predictions}
\end{figure*}
We also evaluated our approach in a more complex visual environment, CoinRun~\cite{cobbe2018quantifying}. This environment consists of procedurally generated platform environments. In particular, the background, player, enemies, platforms, obstacles, and goal locations are all randomly instantiated. The agent can take actions left, right, jump, and down, jump-left, jump-right, and do-nothing. The game ends when the agent reaches a single coin in the game. We used 1000 episodes of expert demonstrations to train ILPO and BCO. 

In these experiments, we evaluated each approach within a single easy and hard level, shown in figure~\ref{fig:thor}. This environment is more difficult than classic control because it uses images as inputs and contains more actions. As such, the dynamics learning takes place over many more dimensions. In particular, the state space consists of $128$x$128$x$3$ pixels and $7$ actions. Thus, ILPO must predict $128$x$128$x$3$ dimensions for the next-state predictions. We found that ILPO performed better when predicting $|Z|=5$ latent actions. This is likely because certain actions, such as moving left, were used less frequently. We use the embedded distance metric from equation~\ref{eq:embedded_distance} to compute the distances between observed and predicted next states.

\subsubsection{Results}
Figure~\ref{fig:thor_results} shows the results for imitation learning. In both the easy and hard tasks, ILPO was not able to perform as well as the expert, but performed significantly better than BCO. As this environment is high-dimensional, it takes more steps to learn the alignment policy than the previous experiments. However, ILPO often learned to solve the task almost immediately, but some random seeds led to bad initialization that resulted in the agent not learning at all. However, good initialization sometimes allows the agent to learn good initial policies zero-shot (see videos in the supplementary for an example). As such, we found that it was possible for the agent to sometimes perform as well as the expert. The results consist of all of the seeds averaged, including those that yielded poor results.

Figure~\ref{fig:predictions} shows the predictions made by the model. ILPO is able to predict moving right and jumping. Because these are the most likely modes in the data, the other generators also predict different velocities. The distance metric is able to correctly select the closest state. 

In general, it can often be difficult to learn dynamics from visual inputs. Unlike BCO, by learning a latent policy first, ILPO is able to reduce the number of environment interactions necessary to learn. BCO requires solving both an inverse dynamics model and behavioral cloning each time it collects a batch of experience. As such, this approach is less efficient than ILPO and in many scenarios would be difficult to perform in realistic environments.

 \section{Discussion and conclusion}
In this paper, we introduced ILPO and described how agents can learn to imitate latent policies from only expert state observations and very few environment interactions. We demonstrated that this approach recovered the expert behavior in four different domains consisting of classic control and vision based tasks. ILPO requires very few environment interactions compared to BCO, a recent dynamics-based imitation from observation approach. In many real world scenarios, unguided exploration in the environment may be very risky but expert observations can be readily made available. Such a method of learning policies directly from observation followed by a small number of action alignment interactions with the environment can be very useful for these types of problems.
 
There are many ways that this work can be extended. First, future work could address two assumptions in the current formulation of the problem: 1) that it requires that actions are discrete and 2) assumes that the state transitions are deterministic. Second, the action remapping step can be made even more efficient by enforcing stronger local consistencies between latent actions and generated predictions across different states. This will drastically reduce the number of samples required to train the action remapping network by decreasing variation between latent and real actions. 

We believe this work will introduce opportunities for learning to observe not only from similar agents, but from other agents with different embodiments whose actions are unknown or do not have a known correspondence. Another contribution would be to learn to transfer across environments.

Finally, our work is complimentary to many of the related approaches discussed. Several algorithms utilize behavioral cloning as a pre-training step for more sophisticated imitation learning approaches. As such, we believe ILPO could also be used for pre-training imitation by observation.

\small
\bibliographystyle{icml2019}
\bibliography{references}

\clearpage
\appendix
\begin{appendices}
\section{Hyperparameters}
We now discuss the hyperparameters and architectures used to train ILPO and behavioral cloning, and BCO. We used the Adam Optimizer to train all experiments. 

\subsection{Classic Control}
\subsubsection{ILPO}
In the latent policy network, the embedding architecture for $E_p$ was: $FC_{128}\rightarrow lrelu\rightarrow FC_{256}$. The leak parameter for $lrelu$ was $.2$ for each case. After converting $z$ to a one-hot, the generator network encodes it with $FC_{256}$, concatenates it with $E_p$ and places it through $lrelu$, then placed it into the following architecture to compute $g(s_t, z): FC_{128}\rightarrow lrelu\rightarrow FC_{dims}$, where dims was the number of state dimensions. To compute $\pi(s|z)$, we compute $lrelu$ for $E_p$ and then the following architecture: $FC_{|A|}$, where $|A|$ is the number of actions. We trained the network for $1000$ epochs with a batch size of $32$. The learning rate was $.0002$.

In the action remapping network, the embedding architecture for $E_a$ was the same as $E_p$. After converting $z$ to a one-hot, the policy network encodes it with $FC_{256}$, concatenates it with $E_a$ and performs $lrelu$, then places it into the following architecture to compute $\pi(a|s_t, z): FC_{64}\rightarrow lrelu\rightarrow FC_{32}\rightarrow FC_{|A|}$. We trained the network for $1000$ steps with a batch size of $32$. The learning rate was $.002$.

\subsubsection{Behavioral cloning}
Behavior cloning used the same architecture as $E_p$ for encoding states, followed by $lrelu$ and then $FC_{|A|}$. 

We trained the network for $1000$ epochs with a batch size of $32$. The learning rate was $.0002$.

\subsubsection{BCO}
BCO used the same architecture as $E_p$ for encoding states then placed it into the following architecture to compute $\pi(a|s_t): FC_{64} \rightarrow lrelu\rightarrow FC_{32}\rightarrow FC_{|A|}$. This is the same architecture as ILPO for $\pi(a|s_t, z)$ except it does not take in latent actions. To compute inverse dynamics for $s_t$ and $s_{t+1}$, we place each through $E_p$ and concatenate them, then compute $lrelu$ for the concatenation and place through $FC_{|A|}$.

We trained the network for $1000$ iterations with a batch size of $32$. At each iteration, we collected $50$ steps of experience from the policy and trained the inverse dynamics model for $10000$ steps and the policy for $10000$ steps. The learning rate was $.0002$. 

ILPO used $\epsilon=.2$ for taking random actions while training and $\epsilon=0$ when evaluating.

\subsection{CoinRun}
\subsubsection{ILPO}
In the latent policy network, the embedding architecture for $E_p$ was: $Conv_{30} \rightarrow lrelu \rightarrow Conv_{60} \rightarrow lrelu \rightarrow Conv_{120} \rightarrow lrelu \rightarrow Conv_{120} \rightarrow Conv_{120} \rightarrow lrelu \rightarrow Conv_{120}$. The parameter for $lrelu$ was $.2$ for each case. After converting $z$ to a one-hot, the generator network encodes it with $FC_{120}$, concatenates it with $E_p$ and does $lrelu$, then places it into the following architecture to compute $g(s_t, z): Deconv_{120} \rightarrow lrelu \rightarrow Deconv_{120} \rightarrow lrelu \rightarrow  Deconv_{120} \rightarrow  lrelu \rightarrow Deconv_{60} \rightarrow  lrelu \rightarrow  Deconv_{30} \rightarrow lrelu \rightarrow Deconv_{15}$. All filters were size $4$x$4$ with stride $2$. We trained the network for $10$ epochs with a batch size of $100$. The learning rate was $.0002$.

In the action alignment network, the embedding architecture for $E_a$ was the same as $E_p$. After converting $z$ to a one-hot, the generator network encodes it with $FC_{256}$, concatenates it with $E_p$ and places it through lrelu, then places it into the following architecture to compute $\pi(a|s_t, z): FC_{64}\rightarrow lrelu\rightarrow FC_{64}\rightarrow FC_{|A|}$. We trained the network for $1000$ steps with a batch size of $100$. The learning rate was $.001$. 

\subsubsection{BCO}
  \begin{figure*}[htb]
 \centering
  \begin{subfigure}{.33\linewidth}
  	\centering
    \includegraphics[width=\linewidth]{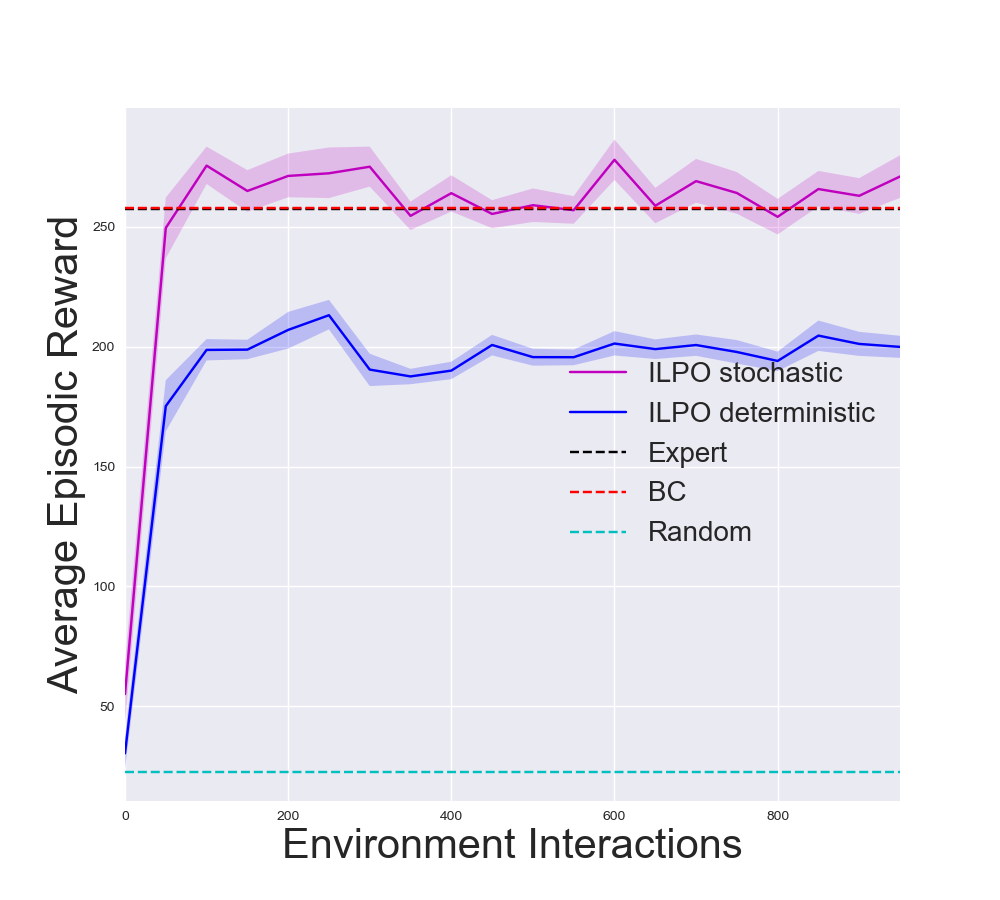}
    \caption{Cartpole}
  \end{subfigure}
    \begin{subfigure}{.33\linewidth}
    \centering
     \includegraphics[width=\linewidth]{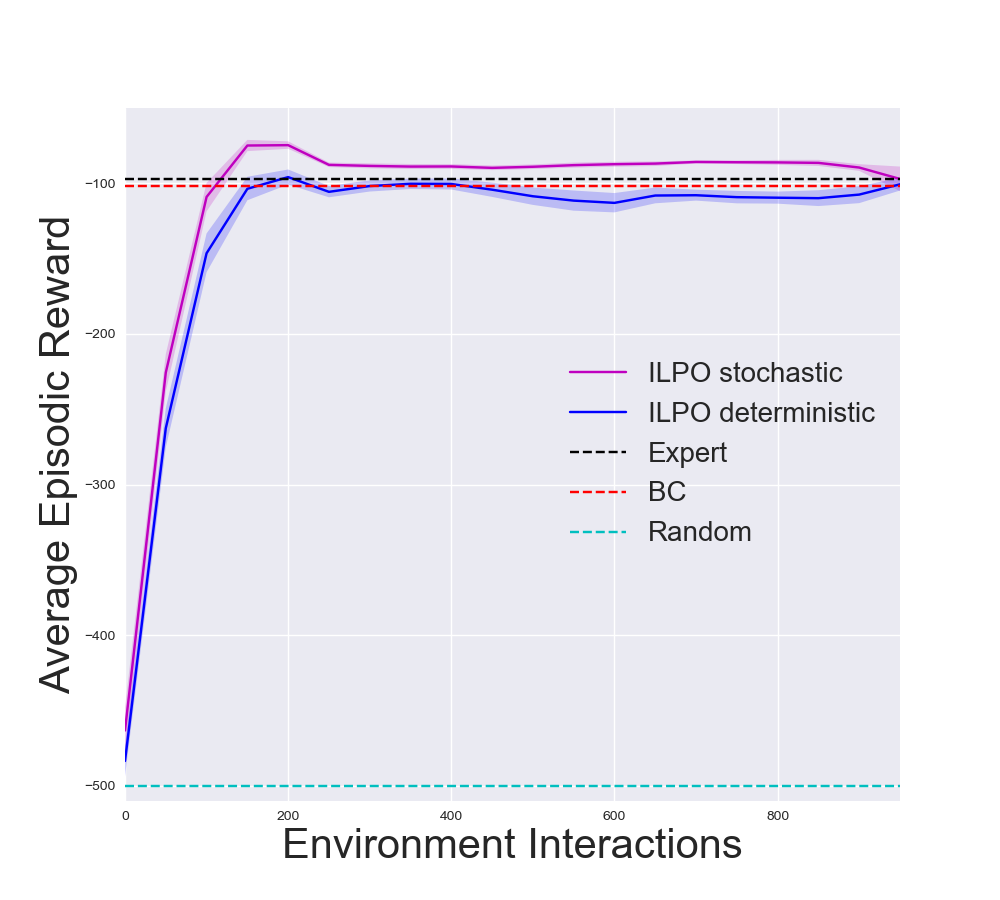}
     \caption{Acrobot}
  \end{subfigure}%
     \caption{Classic control imitation learning results for deterministic and stochastic demonstrations. The trials were averaged over $50$ runs for ILPO and the policy was evaluated every $50$ steps and averaged out of $10$ policy runs. The reward used for training the expert and evaluation was +1 for every step that the pole was upright in cartpole, and a -1 step cost for acrobot.}
\label{fig:stochastic}
\end{figure*}
We found that BCO overfit when it used the same architecture as $E_p$ for encoding states. As such, we first created the state embedding using the following architecture: $Conv_{15} \rightarrow lrelu \rightarrow Conv_{30}$. The parameter for $lrelu$ was $.2$ for each case. All filters were size $4$x$4$ with stride $2$.

We placed the embedding into the following architecture to compute $\pi(a|s_t): FC_{64} \rightarrow lrelu\rightarrow FC_{64}\rightarrow FC_{|A|}$. This is the same architecture as ILPO for $\pi(a|s_t, z)$ except it does not take in latent actions. To compute inverse dynamics for $s_t$ and $s_{t+1}$, we place each through $E_p$ and concatenate them, then compute $lrelu$ for the concatenation and place through $FC_{|A|}$.

We trained the network for $1000$ iterations with a batch size of $100$. At each iteration, we collected $200$ steps of experience from the policy and trained the inverse dynamics model for $500$ steps and the policy for $500$ steps. The learning rate was $.0001$. Training this model was very slow because it was done on images and needed to train the policy and inverse dynamics model at each iteration. 

Both BCO and ILPO used $\epsilon=.2$ for taking random actions while training and $\epsilon=.1$ while evaluating.

\section{Deterministic Demonstrations}
Figure~\ref{fig:stochastic} shows results for using deterministic demonstrations in ILPO. It is clear that stochastic demonstrations are necessary for achieving optimal performance for ILPO, although it still received high reward for both tasks with deterministic demonstrations. Regardless, demonstrations may be stochastic for many reasons---mistakes made by the demonstrator, missed keystrokes, lag, variations in preference, etc. We believe this setting is quite likely and hence not too much of a limitation of our approach.
\end{appendices}
 \end{document}